# An Efficient Inductive Unsupervised Semantic Tagger

K T Lua
Department of Information Systems and Computer Science
National University of Singapore
Lower Kent Ridge Road, Singapore 119260
luakt@iscs.nus.sg

**Abstract**

*We report our development of a simple but fast and efficient inductive unsupervised semantic tagger for Chinese words. A POS hand-tagged corpus of 348,000 words is used. The corpus is being tagged in two steps. First, possible semantic tags are selected from a semantic dictionary(Tong Yi Ci Ci Lin), the POS and the conditional probability of semantic from POS, i.e., $P(S|P)$. The final semantic tag is then assigned by considering the semantic tags before and after the current word and the semantic-word conditional probability $P(S|W)$ derived from the first step. Semantic bigram probabilities $P(S|S)$ are used in the second step. Final manual checking shows that this simple but efficient algorithm has a hit rate of 91%. The tagger tags 142 words per second, using a 120 MHz Pentium running FOXPRO. It runs about 2.3 times faster than a Viterbi tagger.*

## 1. Introduction

Word Sense Disambiguation or WSD has been an important research area in NLP for many years (Black 1988, Bruce 1994 and 1995, Harder 1993, Lam 1995, Leacock 1993, Luk, 1995, Ng 1995, Ng 1996, McRoy 1992, Miller 1994, Yarowsky 1995, Zernik 1990). The reported accuracy of disambiguation varies from 72% (Black 1988) to 90%(Ng 1996). In the Chinese language, unfortunately, there have been much fewer reports on WSD.

Lam(1995) applied a linguistic-based word sense disambiguation algorithm for Chinese(LSD-C). This system does not require training. It relies on two dictionaries: Xiandai Hanyu Cidian and TongYiCi Lilin(Mei, 1983 and 1992) and achieves hit rates of from 36% to 57.6% (with the average of 45.60%). The hit rates seem a bit on the low side. However, the statistics covers only ambiguous words. The test sample has average 3.4 senses per word.

Traditionally, part-of-speech plays a major role in the analysis of sentences. In Western languages, the functional role a word plays in a sentence is almost entirely determined by its part-of-speech, or syntactical category. In Chinese, on the contrary, it is almost impossible to establish a one-to-one association between the part-of-speech of a word and its functional role(Wan 1989, Wu 1982, Zhang, 1986). For example, in Chinese, a noun can be a subject, predicate, object, and attributive(See Table1, adapted from Zhang 1986, page 155). A noun is only not allowed to play the role of a complement.

**Table 1: Functional Roles of Chinese Part-of-Speech**

|  | subject | predicate | object | adjective | adveribal | complement | independent |
|---|---|---|---|---|---|---|---|
| Noun | O | ? | O | √ |  | X | √ |
| Verb | ? | O | ? | √ |  | √ | √ |
| Adjective | ? | O | ? | O | O | √ | √ |
| Quantitative | √ | ? | √ | √ | √ | √ | √ |
| pronoun | O |  | O | √ | √ | ? | √ |
| adverb | X | X | X | X | O |  | √ |

symbols: O frequently allowed, √ : allowed, ? conditionally allowed,  allowed only in a few cases, X: not allowed.

This creates a problem for the Chinese sentence analysis using part-of-speech. In a standard text book of Chinese language, sentence structures are analysed according to the roles of the words in the sentences. It is therefore important to find out if the semantic class(or sense) of a Chinese word plays a key role in analyzing Chinese sentences. We therefore have to have a Chinese text that is semantically tagged.

With the absence of a semantically tagged corpus, we have to use an unsupervised approach. To make this possible, we adopt two important strategies: 1. Induction and 2. Divide-and -conquer.

Using the first strategy, we hand-tagged a small section of the corpus of about 17,000 words. From this corpus we calculate P(S|P), P(S|W) and P(S|S), which are the conditional probabilities of semantic(S) from part-of-speech(P), semantic(S) from word(W) and semantic bi-grams. We use these parameters to guide us in the subsequently tagging of larger and larger corpora. At the end, a corpus of 348,000 words are tagged.

For the second strategy, we divide the tagging into two phases. First, we identify the possible tags and second, we compute for the most likely tag from a list of possible tags. We make our preliminary selection of possible semantic tags by referring to a semantic classification dictionary, i.e., Tong Yi Ci Ci Lin (CILIN, Mei, 1983, 1992) and the conditional probabilities of semantic from part-of-speech, i.e., P(S|P). Final selection of the most likely tags bases on P(S|W) and P(S|S) probabilities.

In this way, we develop a fast and efficient algorithm to semantically tag a Chinese corpus of 348,000 words to an accuracy of about 90%. The tagging algorithm also runs 2.3 times faster that the Viterbi algorithm, one of the fastest tagging algorithm available.

After this introduction, in Section 2, we provide a brief description on our corpus, the part-of-speech tag set and the semantic classification of CILIN. In Section 3, we explain in details our tagging algorithms. In Section 4, we report the steps of tagging. In Section 5, we present a simple error analysis and compare the speed of our tagging algorithm with some those using other approaches, such as genetic algorithm and Viterbi algorithm. Our final conclusion appears in Section 6.

## 2. Our corpus and CILIN's Semantic Classes

We obtained a POS-tagged corpus from Tsing Hua University. This corpus is manually tagged with a tag set of 113 part-of-speech (Bai 1992, 1995, also see Table 2)

We extract a section of text of about 17,000 words from the corpus and manually tag it with semantic classes according to the Tong Yi Ci Ci Lin(Mei, 1983, 1992). CILIN' s semantic classification is a three layers hierarchical tree. There are 12 major, 95 middle and 1428. minor classes(Lua 1993a and 1993b). We select the middle class(95 classes) and add in the following additional classes :

**Ma** numbers
**Nd** name of place
**Nr** name of person, including surnames
**Pd** punctuation marks
**Ud** Others

So, we end up having totally 100 semantic classes.

We select the middle classes as it matches well with the 113 POS tags of the Tsing Hua system. During hand tagging, we select the most appropriate class and assign it to the word according to its meaning and POS in the sentence. In some cases, we have to manually provide a tag for the word. These are: (1) when we have decided that none of the classes in CILIN is appropriate and (2) when the word is absent from CILIN. In case (2), we refer to a word with similar or closest meaning to the one in CILIN. For example, we refer to 是 for semantic classes of 就是. Like any other hand tagged corpora, we cannot ensure that our tagging is 100% correct. However, as we can later, our system has a very high tolerance to errors. There is actually no need to start with a 100% correctly tagged text.

From this hand tagged text, we derive our first set of conditional probabilities. These are : P(S|P), P(S|W) and P(S|S). P(S|P) is the most useful as we rely on it to select the preliminary set of semantic tags for the word.

## 3 Development of tagging algorithm with a small hand-tagged corpus(A)

We develop our tagging algorithm with the above-mentioned hand-tagged corpus(A). Our approach is *inductive* because we use parameters derived from a small section of the corpus to tag a larger and larger section of the corpus. We repeat the process until the whole corpus is tagged.

Table 2: Tsing Hua POS Tag Set

| | | | | | | | | |
|---|---|---|---|---|---|---|---|---|
| nf | {nf} | 姓氏 | qnk | {qn} | 种类量词 | y | {y} | 语气词 |
| npf | {np} | 人名 | qng | {qn} | 名量词"个" | o | {o} | 象声词 |
| npu | {np} | 机构名 | qnm | {qn} | 度量词 | e | {e} | 叹词 |
| npr | {np} | 其它专名 | qns | {qn} | 不定量词 | h | {h} | 名词前缀 |
| ng | {ng} | 普通名词 | qnv | {qn} | 容器量词 | kn | {k} | 名词后缀 |
| t | {t} | 时间词 | qnf | {qn} | 成形量词 | kv | {k} | 动词后缀 |
| s | {s} | 处所词 | qnz | {qn} | 准量词 | i | {i} | 成语 |
| f | {f} | 方位词 | qv | {qv} | 动量词 | j | {j} | 简称语 |
| vg | {vg} | 一般动词 | qt | {qt} | 时量词 | l | {l} | 习用语 |
| vgo | {vg} | 动词不带宾 | rm | {rm} | 代词"每" | x | {x} | 其他 |
| vgn | {vg} | 动词带体宾 | rn | {rn} | 体词性代词 | xch | {x} | 非汉字 |
| vgv | {vg} | 动词带动宾 | rp | {rp} | 谓词性代词 | xfl | {x} | 数学公式 |
| vga | {vg} | 动词带形宾 | rd | {rd} | 副词性代词 | ( | {w1} | |
| vgs | {vg} | 带小句宾 | pg | {p} | 普通介词 | ) | {w1} | |
| vgd | {vg} | 动词带双宾 | pba | {p} | 词介把(将) | " | {w1} | |
| vgj | {vg} | 带兼语宾 | pbei | {p} | 被(让,叫) | " | {w1} | |
| va | {va} | 助动词 | pzai | {p} | 介词"在" | 〔 | {w1} | |
| vc | {vc} | 补语动词 | db | {db} | 副词"并" | 〕 | {w1} | |
| vi | {vi} | 系动词 | dd | {dd} | 程度副词 | 《 | {w1} | |
| vy | {vy} | 动词"是" | dr | {dr} | 一般副词 | 》 | {w1} | |
| vh | {vh} | 动词"有" | cf | {cf} | 前置连词 | 〈 | {w1} | |
| vv | {vv} | 来、去连谓 | cmw | {cm} | 中置(词语) | 〉 | {w1} | |
| vf | {vf} | 形式动词 | cmc | {cm} | 中置(分句) | …… | {w1} | |
| ag | {ag} | 一般形容词 | cms | {cm} | 中置(句子) | ' | {w1} | |
| ac | {ac} | 补语形容词 | cbw | {cb} | 后置(词语) | , | {w1} | |
| z | {z} | 状态词 | cbc | {cb} | 后置(词语) | —— | {w1} | |
| b | {b} | 区别词 | cbs | {cb} | 后置(词语) | / | {w1} | |
| mx | {m} | 系数词 | usde | {usde} | 助词"的" | ~ | {w1} | |
| mw | {m} | 位数词 | uszh | {uszh} | 助词"之" | · | {w1} | |
| mg | {m} | 概数词 | ussi | {ussi} | 助词"似的" | 、 | {w1} | |
| mm | {m} | 数量词 | usdi | {usdi} | 助词"地" | , | {w2} | |
| mh | {m} | 数词"半" | usdf | {usdf} | 助词"得" | 。 | {w2} | |
| mo | {m} | 数词"零" | ussu | {ussu} | 助词"所" | ; | {w2} | |
| maf | {m} | 前助数词 | ussb | {ussb} | 助词"不" | : | {w2} | |
| mam | {m} | 中助数词 | utl | {ut} | 助词"了" | ! | {w2} | |
| mab | {m} | 后助数词 | utz | {ut} | 助词"着" | ? | {w2} | |
| qni | {qn} | 个体量词 | utg | {ut} | 助词"过" | @ | {w2}• | |
| qnc | {qn} | 集合量词 | ur | {ur} | 其它助词 | | | |

## 3. 1 Selection of Possible Semantic Tags - the **PICK** program

We select the 7 most likely tags using CILIN and P(S|W). We set up a score system as below: For every word, we assign a score of 1 to the semantic classes that appear in CILIN. We add this number to P(S|P) according to the POS assigned to the word. We then select from all the 100 semantic classes, the 7 classes with the highest scores. We place the tag with the highest score in the first cell and name it SEM1; the tag with the second highest score in SEM2 and so on. We found that 73.47% of the words have their semantic tags assigned by a combined score from CILIN and P(S|P). For the remaining 26.53% words, the assignments are determined by P(S|P) alone. Note that $0 \leq 0 \leq P(S|P) \leq 1$ and the total score is $\leq 2$.

Table 3: Semantic Classes of CILIN

| | | | |
|---|---|---|---|
| Aa 泛称 | Bl 排泄分泌 | Ec 色味 | Ia 自然现象 |
| Ab 男女老少 | Bm 材料 | Ed 性质 | Ib 生理现象 |
| Ac 体态 | Bn 建筑物 | Ee 德才 | Ic 表情 |
| Ad 籍属 | Bo 机具 | Ef 境况 | Id 物体状况 |
| Ae 职业 | Bp 用品 | Fa 上肢动作 | Ie 事态 |
| Af 身份 | Bq 衣物 | Fb 下肢动作 | If 境遇 |
| Ag 人状况 | Br 食药毒品 | Fc 头部动作 | Ig 始末 |
| Ah 亲属 | Ca 时间 | Fd 全身动作 | Ih 变化 |
| Ai 辈次 | Cb 空间 | Ga 心理状况 | Ja 联系 |
| Aj 人关系 | Da 事情情况 | Gb 心理活动 | Jb 异同 |
| Ak 品行 | Db 事理 | Gc 能愿 | Jc 配合 |
| Al 才识 | Dc 外貌 | Ha 政治活动 | Jd 存在 |
| Am 信仰 | Dd 性能 | Hb 军事活动 | Je 影响 |
| An 丑类 | De 性格才能 | Hc 行政管理 | Ka 疏状 |
| Ba 物体 | Df 意识 | Hd 生产 | Kb 中介 |
| Bb 拟状物 | Dg 比喻物 | He 经济活动 | Kc 连接 |
| Bc 物体部分 | Dh 臆想 | Hf 交通运输 | Kd 辅助 |
| Bd 天体 | Di 社会政法 | Hg 教卫活动 | Ke 呼叹 |
| Be 地貌 | Dj 经济 | Hh 文教活动 | Kf 拟声 |
| Bf 气象 | Dk 文教 | Hi 社交 | La 敬语 |
| Bg 自然物 | Dl 疾病 | Hj 生活 | Nd 地名 |
| Bh 植物 | Dm 机构 | Hk 宗教活动 | Nr 人名 |
| Bi 动物 | Dn 数量单位 | Hl 迷信活动 | Pd 标点 |
| Bj 微生物 | Ea 外形 | Hm 公安司法 | Ma 数字 |
| Bk 全身 | Eb 表象 | Hn 恶行 | Ud 其他 |

### 3.1.1 Hand-tagged Corpus A - 17,000 words

To determine the accuracy of the preliminary selection, we run **PICK** with the hand tagged corpus A. The hit rates vary from 78.69% to 99.87%, depends on the number of classes include(see Table 4)

Table 4: Hit Rate of Preliminary Assignment

| No of tags | Hit Rate |
|---|---|
| 1 | 78.69% |
| 2 | 95.47% |
| 3 | 98.46% |
| 4 | 99.83% |
| 5 | 99.85% |
| 6 | 99.87% |
| 7 | 99.87% |

As the speed of tagging depends on the number of tags included in this preliminary selection(See Table 5), we decide to limit the number of tags for the final selection to 3. This sets an upper limit for our tagging accuracy to 98.46%, a number that we consider to be much higher than what we would achieve from the current tagging algorithm.

### 3.2 . Tagging Algorithm - **TAG** program

In the second step, we compute for the most likely tag from a pool of tags selected from **PICK**. In this program, we consider only the tags before and after the current one. We calcite 18 scores. These are the conditional probabilities of the semantic bigrams(P(S|S)) between the current selection and its 6 neighbors(See Figure 1) weighted by P(S|W) for the word W under consideration

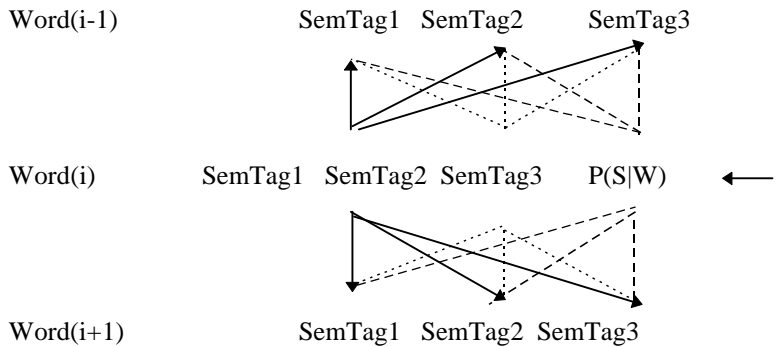

Figure 1: Computation of Score

score = $P(S_i|W_i)[\Sigma P(S_i(W_i)|S_j(W_{i-1})) + \Sigma P(S_i(W_i)|S_j(W_{i+1}))]$
where i, j =1,2,3

In Figure 1, Word(i) is the current word. $Word_{i-1}$ and $Word_{i+1}$ are words before and after the current word. The *weight* between two semantic tags equal to the bigram conditional probabilities between them. These are: P(S|S), or $P(S_i|_{i-1})$ and $P(S_{i-1}|_i)$.

Processing corpus (A) with this algorithm and with the preliminary data set of P(S|W), P(S|S) derived from the hand-tagged corpus, we obtain a hit rate of 86.4%. We consider this high enough for us to begin our inductive process.

### 3.3 Repeated Tagging

Before going to the larger corpus, we want to know if we can repeatedly tag the corpus to obtain better hit rates. We update data (P(P|W), P(S|W) and P(S|S)) with the computer-tagged corpus A and re-tag the corpus. The hit rate increases slightly to 87.4 %. The improvement is thus rather limited.

### 3.4 Probabilities of Items Which Are Absent (Sparse Data Set Problem)

One important decision to make during tagging is the choice of probabilities of occurrence of item which does not occur (sparse data set problem). Although our algorithms allows 0 probability, we believe that it will be more better if we select a number that is slight less than half, i.e. 0.4. We set the probability of non-occurrence items to 0.4/(corpus size).

We experiment our tagging programs with many different values. These are : 0.1, 0.2, 0.3, 0.4 and 1.0. We find that the smaller the value, the more the system selects a tag based on the current parameters. This is undesirable as we want our algorithm to play a major role in the search of correct tags. We also do not want to leave the selection of tag entirely to the preliminary data set. Table 5 shows the hit rates for different values of occurrence of non-occurrence items.

Table 5: Hit Rates and Occurrence of Non-occurrence Items

| No of tags | 2 | 3 | 5 | 7 |
|---|---|---|---|---|
| processing time | 7ms | 20ms | 18ms | 28ms |
| 0..1 | 89% | 93% | 96% | 96% |
| 0.2 | 88% | 92% | 95% | 94% |
| 0.4 | 85% | 86% | 87% | 85% |
| 1.0 | 76% | 75% | 73% | 58% |

## 4. Tagging

With the parameters extracted from the small hand-tagged corpus, we proceed to do preliminary selection of the semantic tags for a larger corpus. We divide this larger corpus into 2 parts, corpus B and corpus C of about equal size, each having about 170,000 words. In this way, we can obtain a better picture on how the tagger performs. Corpus B was processed first.

### 4.1 Tagging Corpora B - 179,159 words

We run **PICK** on Corpus B. Tags of 73.47% of the words are selected by CILIN and P(S|P), while the remaining 26.53% are selected by P(S|P) alone. However, before exciting TAG program, we have to solve a problem. Corpus B has not been tagged before and therefore the parameters P(S|W) are lacking. We do not wish to use P(S|W) derived from Corpus A as Corpus B is 10 times larger and it contains far more number of unique words. Corpus A has about 2300 unique words while Corpus B has more than 10,000 unique words.

We decide do a little 'repair' to CILIN. We update the semantic classes in CILIN by referencing to P(S|W) data obtained from Corpus A. We find that 94.90% of the tags are now selected by the updated CILIN and P(S|P) and only 5.10% are selected by P(S|P) alone.

Next, we use the tags in SEM1 as the preliminary semantic classes for computing P(S|W) and P(S|S). We know that SEM1 contains only 78.68% of the correct tags(See Table 4).

In the next round, we run **TAG** to select the most likely tags. Then we update parameters P(S|P), P(S|S) and P(S|W) and re-tag the corpus with **TAG**. We re-tag the Corpus only once as we find that the two tag sets differ by only 2% (98% in agreement).

### 4.2 Tagging of Corpus C - 167,234 words

For corpus C, the PICK program selects 71.56% of tags by referring to CILIN (original CILIN, not the updated one) and P(S|P) and the remaining 28.44% from P(S|P) alone. We again update CILIN with P(S|W) generated from Corpus B(no more from Corpus A). The corresponding rates now change to 92.92% and 7.09%. Subsequent tagging process are identical to what we have described in Section 4.1.

### 4.3 Tagging of Whole Corpus of 348,393 words

At the final stage, Corpus B and C are combined into a single corpus and the whole corpus is tagged. This is called Corpus Z.

## 5. Results

2000 semantic tags from corpus Z are selected and checked manually. A total of 197 errors are discovered. This gives our tagging a hit rate of 90.1%. It is difficult to present a detailed analysis on the error pattern based on this small error sample. However, we can still classify them into the following types:
1. Errors caused by wrong or inadequate CILIN classification
2. Error caused by wrong POS tagging
3. Errors caused by the tagger

### 5.1 Errors

CILIN has many types of errors. These are: (1) Errors due to the wrong classification by the authors. The 70,000 words are classified and collated manually by the 4 authors in a span of about 10 years. They are many occasions of in-consistency and wrong entries to the dictionary. This is event more serious as we obtain the entries not from the main text, but from the indices where the authors were paying much lesser attention to in their checking and verification.

The second problem is our own entries to the CILIN data base from the dictionary to computer. We have discovered an error of about 1~2% in the data entry. We have not corrected these erroneous entries because of the huge number of words. The last and more serious problem is the inadequacy of the word entries in the dictionary. Many words commonly used today are not included in the dictionary.

11% of the total errors are identified to be caused by the wrong semantic entries to CILIN. The incompleteness of CILIN produces another 8 errors (4%). These are 轴承(5 errors), 法兰盘 ( 1 error), 电镀(1 error), 切割机( 1 error).

The POS tagging of the corpus is also not 100% correct. For example, all idioms are classified as 'i'. This produces 2 errors. We had also contributed 6 errors by not considering class 's' as names of places, for example, 上海(Shanghai).

We may conclude that about 19 errors come from the CILIN and 8 errors from POS and our preliminary semantic tagging. We would have our hit rate improved by 0.95% if these errors are eliminated.

## 5.2 Number of Semantic Tags per Word

The number of semantic tags associated with each word is an important factor to look at. From table 6, we find that the average number of semantic tags per word is 1.11. Compared this number to the 1.20 semantic classes per word obtained by direct counting from CILIN(Lua 1993a, 1993b), we find that the Chinese words are less ambiguous when they appear in text.

Table 6 Semantic Classes per Word

| No. Semantic Tags | Occurrence | Percentage |
| --- | --- | --- |
| 1 | 17008 | 91.21% |
| 2 | 1290 | 6.92% |
| 3 | 210 | 1.13% |
| 4 | 48 | 0.26% |
| 5 | 5 | 0.03% |
| 6 | 12 | 0.06% |
| 7 | 7 | 0.04% |
| 8 | 0 | 0 |
| 9 | 18 | 0.10 |
| 10 | 0 | 0 |
| 11 | 11 | 0.06% |

## 5.3 Distribution of Semantic Classes

We can also compare the dynamic and static distribution of the semantic classes. (See Table 7). It is quite interesting to observe that the two distributions agree to each other very well(Lua 1993a and 1993b). The only exception are classes B and K. For class B, there is a much larger collection in CILIN than it is actually used. Also, for class K, although the number of words in CILIN is quite small, their actual usage is very high. K words are *functional* words that are grammar markers in a sentence.

Table 7: Distribution of Semantic Classes

| Class | Dynamic (from corpus) | Static (from CILIN) |
|---|---|---|
| A | 1.60% | 8.24% |
| B | 2.22% | 19.24% |
| C | 4.31% | 3.94% |
| D | 26.14% | 16.80% |
| E | 7.16% | 15.52% |
| F | 7.27% | 2.70% |
| G | 2.76% | 4.19% |
| H | 10.96% | 16.24% |
| I | 3.06% | 7.59% |
| J | 4.70% | 2.21% |
| K | 20.26% | 3.16% |
| L | 0.001% | 0.19% |

## 5.4 Number of semantic classes per POS class

The number of semantic classes associated with one POS class is an important indicator for us to evaluate the usefulness of the semantic tagging. The semantic tags will be redundant if there is a one-to-one association between the two tag sets. The statistics is given in table 8

Table 8: No of semantic tags associated to each POS tag

| No of Sem | No of POS | No of Sem | No of POS | No of Semantic classes | No of POS |
|---|---|---|---|---|---|
| 1 | 1 | 12 | 1 | 28 | 1 |
| 2 | 32 | 13 | 1 | 31 | 1 |
| 3 | 13 | 15 | 2 | 33 | 1 |
| 4 | 9 | 16 | 2 | 34 | 1 |
| 5 | 5 | 19 | 1 | 38 | 1 |
| 6 | 2 | 20 | 2 | 52 | 1 |
| 7 | 1 | 22 | 1 | 55 | 1 |
| 8 | 2 | 24 | 1 | 60 | 1 |
| 10 | 1 | 25 | 1 | | |
| 11 | 5 | 27 | 1 | | |

It can be seen from the table that most POSs are associated with 2-5 semantic tags. It is therefore quite clear that semantic tags provide information in addition to those provided by POS.

## 5.4. Comparison with Genetic and Viterbi Algorithms

We consider 2 other alternative ways of tagging before we work on the current approach. These are :(1) Genetic Algorithm(GA Tagger), (2) Viterbi Algorithm.

## 5.4.1 GA Tagger

In the GA tagger that we developed to tag the corpus, we use P(P|P), P(P|W), P(S|P), P(S|W), P(S|S) to compute the fitness function. This allows us to tag both POS and semantic classes simultaneously with and without the preliminary tag selection described in Section 3. For Corpus A, we have 100% hit rates. This is because the GA, with all its parameters, memorize the complete tag set. This type of tagging is meaningless.

We next attempt the *outside* test. This is done by removing a sentence from the corpus and re-compute all the probabilities. We then use the new set of parameter to tag the sentence. Experimenting this approach with 20 sentences selected for outside test, the average hit rates are : 80.2% for semantic and

79.7% for POS tagging. The average tagging speed is 48 s per word, using a 120 MHz Pentium PC running Visual FOXPRO Ver 3.0.

To reduce the long tagging time, we pre-selection 3 tags using PICK program and then tag the sentences with the GA tagger. The processing speed increases to 40 ms per word. We eventually abandoned this algorithm as the GA approach is the slowest amongst all the three.

### 5.4.2 Viterbi Tagger

Viterbi first appears to be a very good alternative to GA for its high speed. It gives the most optimal solution. However, to our surprise, it is slower than the simple approach we developed for this project. The Viterbi spends 16 ms to tag a word while our simple algorithm spends less that half of this ammoniate, i.e., 7 ms per word. A comparison chart appears in Table 9. Note that GA and Viterbi select the best tag from a pool of 100 (POS or semantic) or 200 (POS and semantic simultaneously) tags whereas our algorithm selects one from a set of 3 tags
.

Table 9: Speed of Tagging

| Tag Set | 3 | 100 | 200 |
|---|---|---|---|
| Current approach | 7 ms | NA | NA |
| Genetic Algorithm | 40 ms | 48 s | 64 s |
| Viterbi | 16 ms | 1.5 s | 20 s |

We finally abandoned both GA and Viterbi because of their longer processing time. Their possible higher hit rates is not considered an advantage as they required a tagged corpus to act as training examples. With the absence of such a corpus, the higher hit rates cannot be materialized.

## 6. Conclusion

Starting from a hand-POS-tagged corpus, we developed a simple inductive unsupervised process assigning semantic tags to a corpus of 348,393 words. The overall hit rate is estimated to be 90.1%. We further analysis and found that 0.95% of errors are caused by the semantic dictionary, POS tagging and the preliminary semantic tagging, the actual performance of our tagger is 91.05%.

We compare our tagger with 2 other types of taggers in processing speed. The current tagger tags 2.3 times faster than the Viterbi tagger, one of the most efficient tagger. With such as high hit rate, we consider our tagging algorithms fast and efficient. Many useful parameters are derived from this project. These are $P(S|P)$, $P(S|W)$, $P(P|W)$, $P(P|P)$ and $P(S|S)$. These parameters can be used as parameters to tag other corpora.

One problem of the current research is the narrow scope of our corpus that it contains only news items. We suspect that parameters derived from this corpus may not be generally applicable to the tagging of other type of text. For example, text on literature works can be significantly differ from the text of news.

In our next project, we will attempt to tag CKIP, a corpus built by ROCLING. This corpus has been POS tagged by hand and it contains a blanched mix of different types of text. It is a far more better corpus from which more reliable parameters about POS and semantic classes can be derived.

## Reference

Bai 1992, Bai, S.H., Xia, Y and Huang C. N., Automatic Part-of-Speech Tagging System of Chinese, Technical Report, Tsing Hua University, Beijing, 1992.

Bai, Shuanhu, 1995, An Integrated Model of Chinese Word Segmentation and Part-of Speech Tagging, In Advanced and Applications on Computational Linguistics, Page 56-61, Third National Computational Linguistics Meeting, 5-7 Nov, 1995, Shanghai.

Black, Ezra, 1988, An experiment in Computational discrimination of English Word Senses, IBM
Journal of Research and Development, 32(2):185-194.

Brill, Eric, 1992, A simple rule-based part-of-speech tagger, Proceeding of the 3$^{rd}$ Conference on Applied Natural Language Processing(ACL), pp 152-155.

Bruce R., 1995, A Statistical Method for Word Sense Disambiguation, Ph.D. Thesis, New Mexico State University.

Bruce, Rebecca and Janyce Wiebe, 1994, Word Sense disambiguation Using Decomposable Models, In Proceedings of 32$^{nd}$ Annual Meeting of Association for Computation Linguistics, Las Cruces, New Mexico.

CKIP - Chinese Knowledge Information Processing Group, Technical Report 95-02, Institute of Information Science, Academia Sinica (Taiwan).

Fogel, Davide, B, 1995, Evolutionary Computation, IEEE Press.

Harder L. B., 1993, Sense Disambiguation Using On-Line Dictionaries, Natural Language Processing: The PLNLP Approach, Kluwer Academic Publishers, 247-261.

Kupiec, Robust, 1992, part-of-speech tagging using a Hidden Markov Model Computer Speech and Language, Vol 6, No 3, pp225-242.

Lam Sze-Sing, Vincent Y. Lum, Kam-Fai Wong, 1995, Determination of Word Sense In Chinese Full Text Using A Standard Dictionary and Thesaurus. Proceedings of the 1995 International Conference on Computer Processing of Oriental Languages, Honolulu, Hawaii, Nov 23-25, 1995, Page 247-250.

Lin, M. Y. and W. H. Tsai, 1987, Removing the ambiguity of phonetic Chinese input by the relaxation technique, Computer Processing of Chinese and Oriental Languages, Vol 3, No 1, May, Pp1-24.

Leacock, Claudia, Geoffrey Towell and Ellen Voorhees, 1993, Corpus-based statistical Sense Resolution, In Proceedings of the ARPA Human Language Technology Workshop.

Lin, Y.C., T. H. Chiang and K.Y. Su, 1992, Discrimination oriented probabilistic tagging, Proceeding of ROCLING V, Pp87-96.

Liu, Shing-Huan, Ken-jiann Chen, Li-ping Chang and Yeh-Hao Chin, 1995, Automatic Part-of-speech tagging for Chinese corpora, Computer Processing of Chinese and Oriental Languages, Vol 9, No 1, pp31-47.

Lua, K. T. 1993a, A Study of Chinese Word Semantics, Computer Processing of Chinese and Oriental Languages, Vol 7, No. 1. P37-60.

Lua, K. T. 1993b, A Study of Chinese Word Semantics and Its Prediction, Computer Processing of Chinese and Oriental Languages, Vol 7, No. 2, P167-180.

Luk, Alpha, K. 1995, Statistical Sense Disambiguation With Relatively Small Corpora Using Dictionary Definition, In Proceedings of the 33$^{rd}$ Annual General Meeting of Association for Computational Linguistics, Cambridge, Massachusetts.

McRoy, Susan W., 1992, Using Multiple Knowledge Sources For Word Sense Discrimination, Computational Linguistics, 18(1):1-30.

Mei 1983, Mei Jiaju, Zhu Yiming, Gao Yunqi and Yin Hongxiang, Tong Yi Ci Ci Lin, Shanghai Dictionaries Publisher, 1983.
梅家驹，竺一鸣，高蕴琦，殷鸿翔，《同义词词林》，上海辞书 出版社， 1983.
Mei (1992), Mei, Jiaju and Gao, Yunqi, A Study of the formalization of semantics, Communications of COLIPS, Vol 2, No 1, Page 40-27, 1992.

Miller, George A., Martin Chodorow, Shari Landes, Claudia Leacock and Robert G. Thomas, 1994, Using A Semantic Concordance For Sense Identification, In Proceeding of the ARPA Human Language Technology Workshop.

Ng, Hwee Tou, 1995, Word Sense Disambiguation Via Exemplar-Based Classification: A Case Study, Proceedings of NUS Inter-Faculty Seminar on Natural Language Processing, 8-9 Sept. 1995, 1-9.

Ng, Hwee Tou and Hian Beng Lee, 1996, Integrating Multiple Knowledge Sources to disambiguation Word Sense: An Exemplar-Based Approach, to appear in ACL-96.

Pong T. Y. and J. S Chang, 1993, A study of word segmentation and tagging for Chinese, Proceedings of ROCLING VI, Pp173-193 (In Chinese)

Yarowsky, David, 1995, Unsupervised Word Sense Disambiguation Rivaling Supervised Methods, In Proceedings of the 33$^{rd}$ Annual Meeting of the Associations for Computational Linguistics, Cambridge, Massachusetts.

Yu 1995, Yu Shiwen, Tagged Singapore Chinese School Texts, Paper R95001, CommCOLIPS, Vol 5 Page 81-86.

Wan, Huzhou, 1989, Comparison of Chinese and English Lexicon, published by Chinese Foreign Economics and Trading Publisher. 万惠洲，《汉英构词法比较》，中国对外经济贸易出版社, 1989.

Wu, Min Jie, 1982, A Handbook of Chinese and English Grammar, published by Zhishi Chubanshi, Zhejiang, China. (In Chinese)
吴敏洁，《汉英语法手册》， 知识出版社， 1982。

Zernik Uri, 1990, Tagging Word Senses in Corpus: The Needle In The Haystach Revisited, Technical Report 90CRD198, GE R&D Center.

Zhang, Song Lin, 1986, Tabulated Grammar For Contemporary Chinese Language, published by Sichua Kexue Jishu Chubanshe (In Chinese). 张松林，《现代汉语语法表解》， 四川科学技术出版社. 1986